\pdfoutput=1

\documentclass[11pt, hyphens, table]{article}

\usepackage{EMNLP2023}

\usepackage{times}
\usepackage{latexsym}

\usepackage[T1]{fontenc}

\usepackage[utf8]{inputenc}

\usepackage{microtype}

\usepackage{inconsolata}

\usepackage{amsfonts}
\usepackage{amsmath}
\usepackage{graphicx}
\usepackage{bm}
\usepackage{listings}
\usepackage{booktabs}
\usepackage{graphics}
\usepackage{etoolbox}
\usepackage{setspace}

\definecolor{darkgreen}{rgb}{0.33, 0.42, 0.18}
\definecolor{purple}{rgb}{0.69, 0.14, 0.26}
\definecolor{darkpastelred}{rgb}{0.76, 0.23, 0.13}
\lstdefinelanguage{LispSchemas}{
  language     = Lisp,
  keywords={header, types, rigid-conds, static-conditions, static-conds, preconditions, pre-conds, postconditions, post-conds, goals, episodes, necessities, certainties},
}
\lstdefinelanguage{DialogueExamples}{
    language = Lisp,
    basicstyle=\scriptsize,
    keywords={},
    otherkeywords={SOPHIE, DAVID, System},
    keywords = [2]{User},
    keywordstyle=\color{purple},
    keywordstyle=[2]\color{blue},
}
\lstdefinelanguage{Prompt}{
    language = Lisp,
    basicstyle=\scriptsize,
    stringstyle=\color{gray},
    moredelim=[s][\color{blue}]{<}{>},
    keywords={},
    otherkeywords={[System], [User], [Assistant]},
    keywords = [2]{},
    keywordstyle=\color{red},
    keywordstyle=[2]\color{blue},
    deletekeywords={Use, Rewrite, Person}
}

\newcommand\doubleplus{+\kern-1.3ex+\kern0.8ex}

\AtBeginEnvironment{quote}{\par\singlespacing\small}

\title{Get the Gist? Using Large Language Models for Few-Shot Decontextualization}

\author{Benjamin Kane \\
  University of Rochester \\
  \texttt{bkane2@ur.rochester.edu} \\\And
  Lenhart Schubert \\
  University of Rochester \\
  \texttt{schubert@cs.rochester.edu} \\}

\begin{document}
\maketitle

\begin{abstract}
In many NLP applications that involve interpreting sentences within a rich context -- for instance, information retrieval systems or dialogue systems -- it is desirable to be able to preserve the sentence in a form that can be readily understood without context, for later reuse -- a process known as ``decontextualization''. While previous work demonstrated that generative Seq2Seq models could effectively perform decontextualization after being fine-tuned on a specific dataset, this approach requires expensive human annotations and may not transfer to other domains. We propose a few-shot method of decontextualization using a large language model, and present preliminary results showing that this method achieves viable performance on multiple domains using only a small set of examples.
\end{abstract}

\section{Introduction}

As large language models (LLMs) improve in capabilities, work in NLP is increasingly turning towards systems that rely on natural text as a core representation, and that use LLMs as a foundation for reasoning \cite{bommasani-2022-foundation, park-2023-generative}. In many cases, this requires the ability to preserve text from an input source in a form that can readily be reused for downstream tasks.

\begin{figure}[t]
    \centering
    \includegraphics[width=\linewidth]{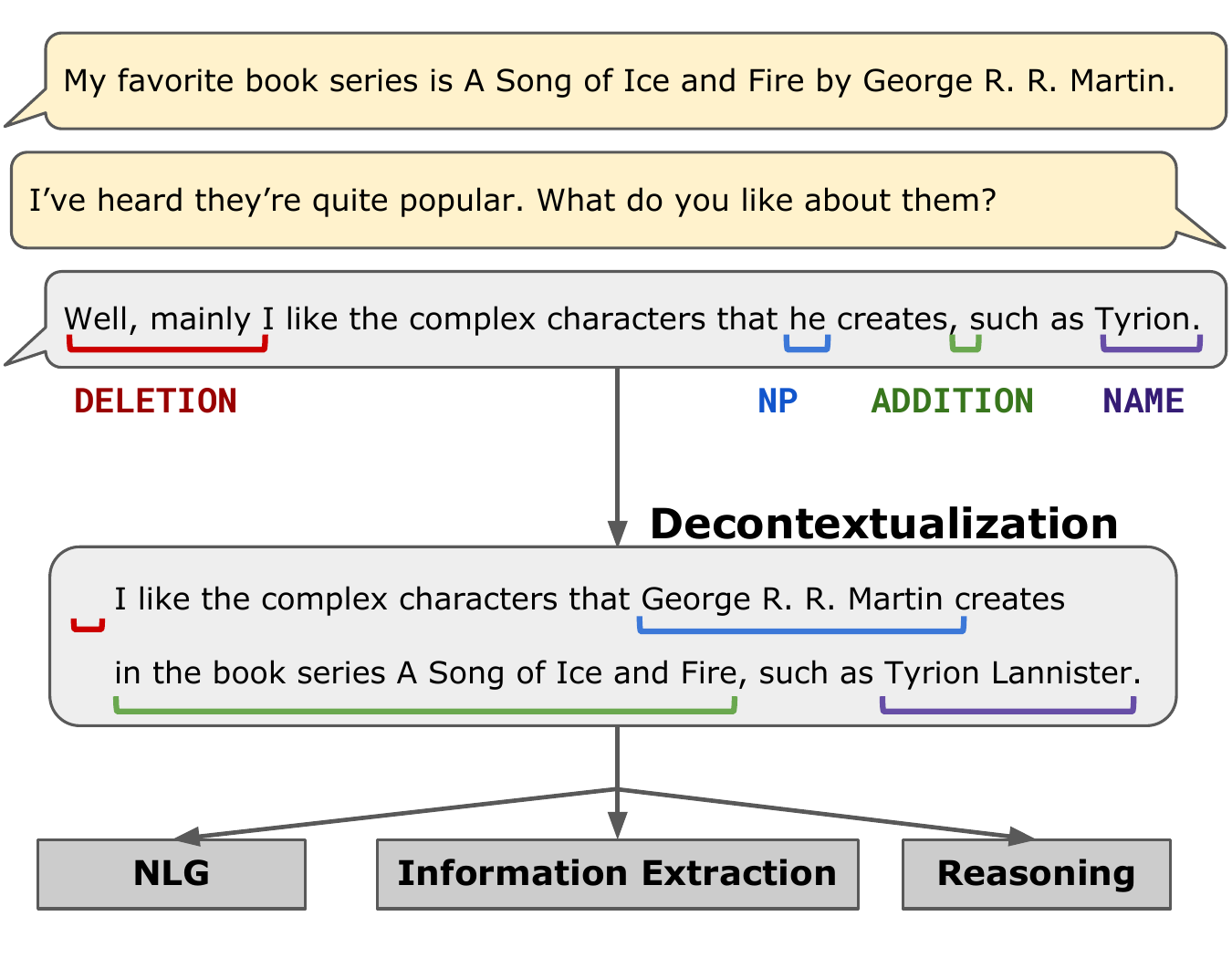}
    \caption{An example of decontextualization within a conversational context (indicated in yellow). The types of edits used to produce the decontextualized sentence -- based on the scheme proposed by \cite{choi-2021-decontextualization} -- are shown below the original and decontextualized sentences.}
    \label{fig:decontextualization}
\end{figure}

However, sentences in natural corpora are often heavily dependent on the surrounding context, making it difficult to extract a sentence directly. Sentences may contain anaphors that refer to other entities in the context, or they may contain discourse markers that relate to the overall structure of the embedding document, or they may contain a variety of elided material. Yet, in many cases it's possible to map a sentence into a \textit{semantically equivalent} form that can be understood in the absence of context. For example, in the conversation in Figure \ref{fig:decontextualization}, given the previous two turns of context, an agent's utterance can be mapped to a form that can be readily understood without context -- allowing it to be reused by downstream tasks.

\citet{choi-2021-decontextualization} provide a formal definition of this task -- known as \textit{decontextualization} -- and decompose the process into several stages of edits. However, this approach -- based on fine-tuning pretrained coreference models and language models -- relied on the crowdsourcing of large numbers of decontextualization annotations, which may be infeasibly expensive, and may not transfer to other domains. A few-shot approach to decontextualization would allow this technique to be more widely adopted in NLP system design.

In this paper, we present a few-shot approach to decontextualization that uses an LLM to map a sentence to a decontextualized form through a series of edits. We present preliminary results involving both automatic and human evaluations demonstrating that this method can achieve viable performance across multiple domains, using only a single small set of annotated examples.

\section{Related Work}

The term ``decontextualization'' was initially introduced by \citet{parikh-2020-totto}, and the concept was further refined and generalized by \citet{choi-2021-decontextualization}. The latter relied on fine-tuning models on large amounts of annotated data. \citet{shin-2021-constrained} demonstrated a \textit{few-shot} method for deriving decontextualized canonical forms by using constrained decoding procedure. However, this approach is only viable within a closed domain where a grammar for the constrained language can be created.

Decontextualization is closely related to, but not identical to, the task of text summarization \cite{gambhir-2017-recent}. In summarization, a sentence need not be rendered into a form that can stand without context; indeed, a summary is often retrieved or generated relative to the context provided by a query. Decontextualization, therefore, is a more constrained problem that involves resolving complex linguistic phenomena such as anaphora and ellipsis that may be ignored by common text summarization methods.

\section{Method}

Given a context $C = \{ c_1, ..., c_n \}$ and a sentence $s$, the goal of decontextualization is to produce a new sentence $s'$ such that $s'$ is interpretable in the empty context, and carries the same truth-conditional meaning as $s$ does given context $C$.

\begin{figure}[t]
    \centering
    \includegraphics[width=\linewidth]{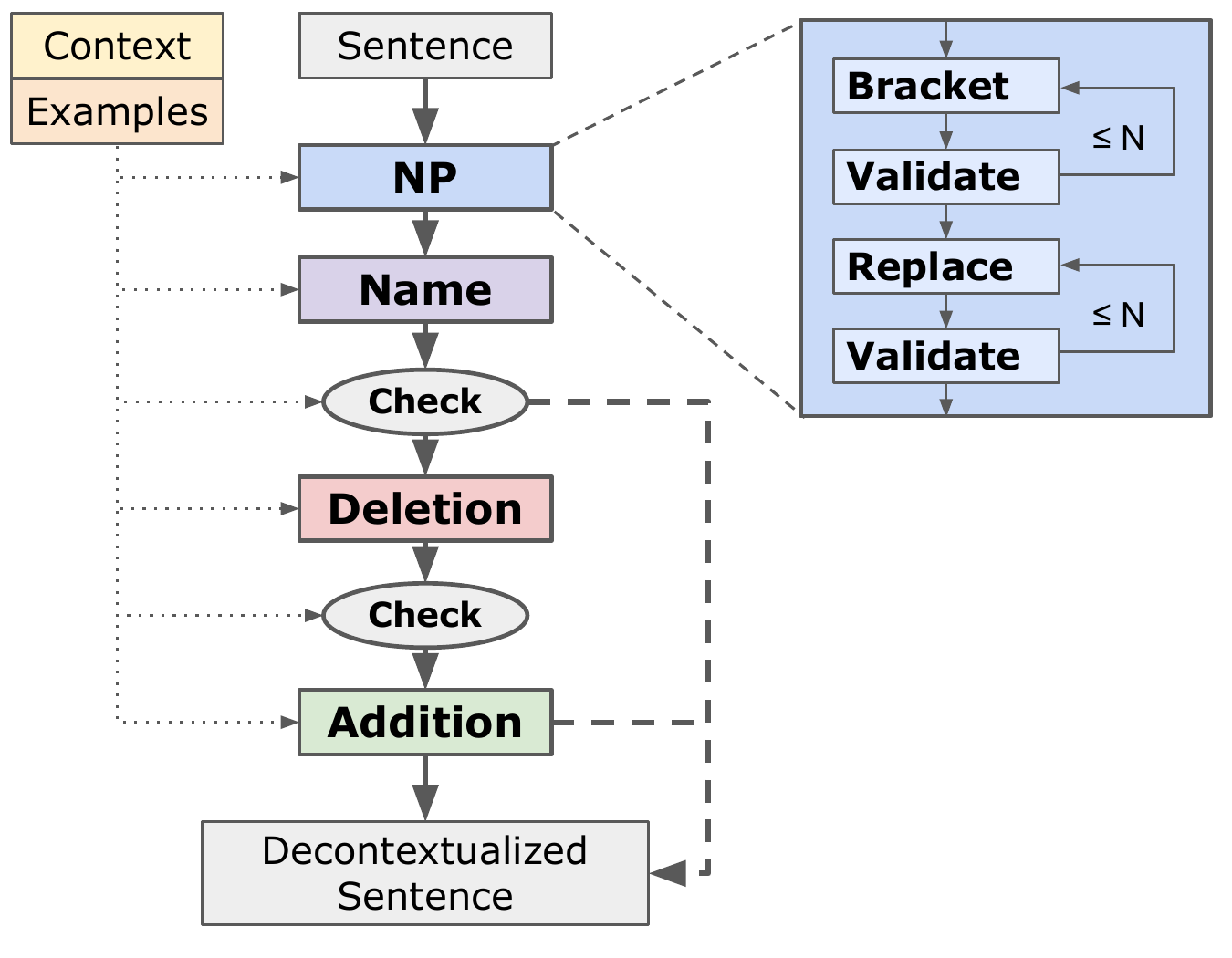}
    \caption{A diagram of our few-shot pipeline. Given an input sentence, context, and set of examples, a sequence of edit nodes transform the input sentence into a decontextualized sentence. Each edit node consists of several substeps to ensure validity, and some edit nodes may also be optionally preceded by cutoff checks.}
    \label{fig:method}
\end{figure}

We propose a pipelined approach to few-shot decontextualization using the \textsc{gpt-3.5-turbo} LLM\footnote{\url{https://platform.openai.com/docs/models/overview}}, based upon the categorization of possible edits described by \citet{choi-2021-decontextualization}. Our method, diagrammed in Figure \ref{fig:method}, consists of several edit nodes that transform a given sentence $s$ sequentially, each provided the context $C$ and a set of examples.

We further decompose each edit node into several substeps (shown for the ``NP'' node in Figure \ref{fig:method})\footnote{We found direct edit prompts to be hallucination-prone.}. To ensure that each substep is accurate, we employ validator functions that compare the input sentence to the output sentence\footnote{The same validator functions are used for each edit node.}; if the LLM cannot generate a correct output after $N$ retries, then the input sentence is returned for that substep. Each edit step has access to $K$ in-context examples for that edit type\footnote{Each example contains a bracketed and an edited sentence, though the former may be derived from the latter.}. We elaborate on each component below; full details and prompts can be found in Appendix \ref{app:method-details}).

\paragraph{Bracket Substep}

We first prompt the LLM to \textit{bracket} each candidate span for that edit type, using ``['' and ``]'' as special delimiter tokens. For example, during the ``NP'' step in Figure \ref{fig:method}, the sentence from Figure \ref{fig:decontextualization} may be bracketed as ``Well, mainly, I like the complex characters that [he] creates, such as Tyrion.''. The validator function for this substep ensures that the output string is identical to the input apart from brackets.

\paragraph{Replace Substep}

Next, the LLM is prompted to \textit{replace} each bracketed expression with edits of the appropriate type. E.g., after bracketing the previous sentence, the expression ``[he]'' may be replaced with ``[George R. R. Martin]''. We validate this substep by ensuring that the non-bracketed sections of the output string match those of the input string. However, we allow for some tolerance by thresholding based on the Jaccard similarity between the uni-grams of the input and output sentence, excluding brackets: $J(S_I, S_O) = \frac{|S_I \cap S_O|}{|S_I \cup S_O|} \geq 0.5$.

\paragraph{Completion Checks}

To avoid overmodification, we allow for cutoff checks -- ``Check'' in Figure \ref{fig:decontextualization} -- to be optionally introduced prior to certain edit nodes; in our pipeline, we use cutoff checks for the ``Deletion'' and ``Addition'' stages. If the LLM classifies a sentence as sufficiently decontextualized (given $K$ examples), the sentence is returned and all subsequent edit nodes are skipped.

\section{Experiments}

\subsection{Datasets}

For our primary experiment, we use the Decontext dataset\footnote{\url{https://github.com/google-research/language/tree/master/language/decontext}} created by \citet{choi-2021-decontextualization}. This dataset, oriented towards text summarization, consists of sentences from the Wikipedia corpus embedded within a context paragraph, each annotated with decontextualized sentences by up to 5 annotators.

We also evaluate whether the performance of our method transfers to a conversational dataset, using the same example set from the Decontext dataset. For this, we use a subset of the validation split from the Switchboard corpus\footnote{\url{https://huggingface.co/datasets/swda}} \cite{godfrey-1992-switchboard}. Additional details about our data preprocessing can be found in Appendix \ref{app:exp-details}.

\subsection{Baselines}

We consider two baselines for each experiment, following \cite{choi-2021-decontextualization}. First, we consider a method \textbf{\textsc{repeat}} that simply repeats the original sentence as an output. Second, we consider a \textbf{\textsc{human}} generated sentence, chosen by taking the median length annotation for each data sample and using the remaining annotations as references.

\subsection{Metrics}

We use the following automatic metrics as our primary method of comparison. First, we report the average percentage increase in length of the decontextualized sentences over the original sentences (\textbf{Len inc.}), as well as the percentage of items where the decontextualized sentence was not identical to the original sentences (\textbf{\% edited}). Next, we report the percentage of items where the decontextualized sentence was an exact match to at least one of the gold annotations, excluding punctuation and stopwords (\textbf{\% match}). Since some items did not require edits, we report this score both for all items, and for the subset of items where all gold annotations contained an edit.

Finally, we report the (\textbf{SARI}) score \cite{xu-2016-optimizing}, which allows us to compute precision/recall/F1 scores between the output and gold annotation for \textit{edits} relative to the input sentence. We compute this metric separately for \textbf{add} and \textbf{delete} edits (micro-averaging over all items), looking at unigrams only and using fractional counts for items with multiple gold annotations.

\subsection{Decontextualization Experiment}
\label{subsec:decontext-experiment}

\begin{table}[t]
\resizebox{\columnwidth}{!}{%
\begin{tabular}{l | c r | r | c c}
    \toprule
    \textbf{Method}  & Len   & \%    & \% match    & SARI add  & SARI del \\
                     & inc.  & edit  & all / edit  & F1 (P/R)  & F1 (P/R) \\
    \midrule
    \textsc{repeat}  & 0   & 0   & 36 / 0   & 0  {\small (0/0)}    & 0  {\small (0/0)} \\
    \textsc{-check}  & 32  & 92  & 8  / 0   & 24 {\small (21/28)}  & 8  {\small (5/40)} \\
    \textsc{+check}  & 24  & 91  & 10 / 5   & 31 {\small (33/29)}  & 21 {\small (15/37)} \\
    \textsc{human}   & 24  & 78  & 44 / 30  & 55 {\small (63/50)}  & 59 {\small (61/58)} \\
    \bottomrule
\end{tabular}%
}
\caption{Automatic evaluation results on the Decontext test set for each set of decontextualizations.}
\label{tab:decontext-automatic-eval}
\end{table}

We first evaluate the performance of our method on a random subset of $1000$ items (approximately 50\% of the data) from the Decontext test split\footnote{We use a subset of the data due to cost considerations.}. We generate $K=20$ in-context examples from the annotations in the development split of the Decontext dataset, filtering for items with 1-2 context sentences and each sentence having less than 30 words. We use an 80/20 ratio of positive to negative\footnote{I.e., examples where no edit is necessary.} examples, and a 50/50 split for cutoff nodes.

\subsubsection{Automatic Evaluation}

We generate decontextualized sentences both with (\textsc{+check}) and without (\textsc{-check}) the cutoff check nodes in Figure \ref{fig:method}. Results for our automated evaluation metrics are shown in Table \ref{tab:decontext-automatic-eval}. We observe that including the cutoff check steps in the pipeline helps avoid extraneous modifications, leading to an average length increase that reflects human annotations, and substantially higher SARI F1 scores.

The scores achieved by our best performing method, while significantly above the baseline, are still well below human-level annotation; Further performance gains can likely be achieved by improved validation and filtering of delete edits (SARI del), which we found have quite low precision relative to recall. However, we note that our method tends to edit a larger fraction of sentences relative to the human annotators, likely diminishing its exact match and SARI add scores despite not being an inherent limitation\footnote{We note that whether more or less information in the decontextualized sentence is desirable may depend on the particular application.}. For this, we turn to a human evaluation of the decontextualized sentences.


\subsubsection{Expert Evaluation}
\label{subsubsec:expert-evaluation}

\begin{table}[t]
\resizebox{\columnwidth}{!}{%
\begin{tabular}{l | r r r | r || r}
    \toprule
             & LLM   & Either   & Human  & Sum  & \% valid  \\ \hline
    LLM      & 6     & 4        & 1      & 11   & 74 \\
    Either   & 6     & 45       & 11     & 62   & -  \\
    Human    & 3     & 6        & 18     & 27   & 87 \\ \hline
    Sum      & 15    & 55       & 30     & 100  &    \\ \hline \hline
    \% valid & 75    & -        & 88     &      &    \\
    \bottomrule
\end{tabular}%
}
\caption{Preferences between the LLM output and human annotations, as well as overall \% marked as valid, with columns/rows showing judgments of expert A/B.}
\label{tab:human-eval}
\end{table}

Due to the wide space of possible ``acceptable'' decontextualized sentences, we also ground our automatic evaluation in an expert evaluation of the generated decontextualizations. We randomly selected 100 examples from our Decontext test subset; given pairs of randomly shuffled candidate decontextualizations, two of the authors annotated each pair for (a) whether each candidate is a valid decontextualization, and (b) which candidate is preferred (allowing for ``either'', i.e., indifference).

Our results are shown in Table \ref{tab:human-eval}. On average, the annotators judged 74.5\% of LLM outputs as being sufficiently decontextualized, vs. 87.5\% of human annotations; interannotator agreements measured by Cohen's kappa were 0.76 and 0.68. The preference annotations indicate that, while human annotations were slightly preferred to LLM outputs, in the majority of cases the expert annotators were \textit{indifferent between the two decontextualizations}.

\subsection{Conversational Transfer Experiment}

One question of interest is whether extending our approach to a new domain or application -- such as extracting ``gist clauses'' in a conversational system \cite{razavi-2017-managing-dialogue-schemas} -- requires a set of new hand-annotated examples, or whether the LLM's performance using the previous set of examples transfers to the new domain. To test this, we replicate the previous automatic evaluation on a small annotated subset of the Switchboard corpus, using the same 20 examples from Section \ref{subsec:decontext-experiment}.

\subsubsection{Annotation Collection}
\label{sec:swda-annotation}

After preprocessing data from the Switchboard validation set, we randomly select 150 sentences that have been determined by an LLM to be interpretable within a 2-turn context window for decontextualization by three expert annotators\footnote{Two of the authors + a PhD student studying NLP.} -- see Appendix \ref{app:exp-details} for more details about this procedure. Annotator agreement, which we compute using mean pairwise Jaccard similarity between annotators over sets of added/deleted unigrams, was significant -- about 0.56 and 0.64 respectively.

\subsubsection{Automatic Evaluation}

\begin{table}[t]
\resizebox{\columnwidth}{!}{%
\begin{tabular}{l | c r | r | c c}
    \toprule
    \textbf{Method}  & Len   & \%    & \% match    & SARI add  & SARI del \\
                     & inc.  & edit  & all / edit  & F1 (P/R)  & F1 (P/R) \\
    \midrule
    \textsc{repeat}  & 0   & 0   & 19 / 0   & 0  {\small (0/0)}    & 0  {\small (0/0)} \\
    \textsc{+check}  & 34  & 96  & 5  / 5   & 27 {\small (22/34)}  & 47 {\small (56/41)} \\
    \textsc{human}   & 8   & 84  & 44 / 37  & 62 {\small (63/61)}  & 75 {\small (77/73)} \\
    \bottomrule
\end{tabular}%
}
\caption{Automatic evaluation results on the Switchboard annotated subset for each set of decontextualizations.}
\label{tab:swda-automatic-eval}
\end{table}

We generate results for the annotated Switchboard data using the best performing method from \ref{subsec:decontext-experiment} (i.e., using cutoff checks); shown as \textsc{+check} in Table \ref{tab:swda-automatic-eval}. As before, the LLM tends to edit at a higher rate than human annotators, leading to a low percentage of exact matches. However, we achieve a comparable F1 score for SARI add as in Section \ref{subsec:decontext-experiment}, and actually achieve a significantly higher F1 score for SARI del -- potentially due to the prominence of discourse markers and other removable content in the Switchboard sentences relative to the Decontext sentences.

\subsection{Qualitative Error Analysis}

Of the items marked as invalid by both annotators in \ref{subsubsec:expert-evaluation}, 15 were due to missing NP edits (typically unresolved pronouns or definite NPs); 4 were due to failures to ADD disambiguating postmodifiers; and 2 were due to a failure to DELETE discourse markers. On inspection of the generated Switchboard decontextualizations, we found that missing DELETE edits were a relatively more common form of error -- likely due to containing forms of discourse markers that were not common in the Decontext example set. Some specific examples for both datasets are shown in Appendix \ref{app:examples}.

\section{Conclusion and Future Work}

We proposed a few-shot LLM-based pipeline for performing decontextualization through a series of edits resembling human annotations, and presented preliminary results showing that this method achieves reasonable performance across multiple domains using few examples. In the future, we believe that the performance of our method can be improved by incorporating smaller specialized NLP models -- e.g., a coreference model -- as well as by experimenting with additional edit types for more complex linguistic phenomena, such as Wh-Question gaps or conversational implicatures.

\section*{Limitations}

Our method is based on a ``pure'' LLM prompting strategy, and achieves lower performance in automatic metrics than the fine-tuned language models explored by \cite{choi-2021-decontextualization}. While our work aims to demonstrate that \textit{viable} results can be achieved in a few-shot setting, in settings where large numbers of annotations are feasible to collect, it is still likely a better option to use a model specifically trained for that task. Additionally, due to cost and resource constraints, we show our results using a GPT-3.5 model; performance may differ using the more recent GPT-4 model.

The automatic metrics used in our paper may be difficult to intuitively interpret, due to the open-ended nature of the possible edits that human annotators can make. While we ground our results for the Decontext dataset in an expert evaluation, our results should still be considered preliminary -- further human evaluations, ablation studies, as well as user studies for downstream tasks using the proposed pipeline are likely necessary to fully assess the utility of this approach.

\section*{Ethics Statement}

Since this paper proposes a heavily constrained pipeline for mapping sentences to a semantically equivalent form (borrowing from a user-provided context), we do not believe that it presents notable ethical concerns in itself. Nevertheless, we would suggest that applications of this method in sensitive domains implement stricter validation functions than those used in this paper, in order to safeguard against potential hallucinated LLM outputs.



\bibliographystyle{acl_natbib}
\bibliography{anthology,refs}

\clearpage
\appendix

\section{Method Details}
\label{app:method-details}

\subsection{Hyperparameters}

We use the \textsc{gpt-3.5-turbo} LLM for all generation. We use the default hyperparameters, i.e., a temperature of 1, top p 1, frequency penalty 0, and presence penalty 0. We use $N=2$ retries for substeps that fail validation, and $K=20$ in-context examples for each step.

\subsection{LLM Prompts}

For each edit node in Figure \ref{fig:method}, we show the LLM prompts that are used for bracketing in Table \ref{tab:prompts-bracket}, and the prompts that are used for replacing in Table \ref{tab:prompts-replace}. For the cutoff checks, we use the following prompt: \textit{``Given a context and a sentence, decide whether the meaning of the sentence can be understood without the context. Answer ``True'' if the sentence can be understood without context, and ``False'' otherwise.''}

\begin{table}[t]
\resizebox{\columnwidth}{!}{%
\begin{tabular}{l | p{\linewidth}}
    \toprule
    NP   & Given a sentence, put brackets around any personal pronouns, definite pronouns, and definite noun phrases that can be replaced with more specific expressions. If there are none, give the original sentence. \\ \hline
    NAME & Given a sentence, put brackets around any acronyms, nominative pronouns, or proper names that can be replaced with more specific expressions. If there are none, give the original sentence. \\ \hline
    DEL  & Given a sentence, put brackets around any discourse markers and connectives that can only be understood in context. If there are none, give the original sentence. \\ \hline
    ADD  & Given a sentence, insert empty brackets wherever additional modifiers should be added in order to allow the sentence to be interpretable without context. If there is no need for modifiers, give the original sentence. \\
    \bottomrule
\end{tabular}%
}
\caption{The LLM prompts that are used for bracketing at each edit node.}
\label{tab:prompts-bracket}
\end{table}

\begin{table}[t]
\resizebox{\columnwidth}{!}{%
\begin{tabular}{l | p{\linewidth}}
    \toprule
    NP   & Given a context and a sentence, replace any bracketed expressions in the sentence with a more explicit referring expression from the context or general knowledge. If there are no bracketed expressions, do nothing. \\ \hline
    NAME & Given a context and a sentence, replace any bracketed expressions in the sentence with a more explicit referring expression from the context or general knowledge. If there are no bracketed expressions, do nothing. \\ \hline
    DEL  & Given a context and a sentence, remove any bracketed expressions if they are extraneous or require context to interpret. If there are no bracketed expressions or if there is no need to make any changes, do nothing. \\ \hline
    ADD  & Given a context and a sentence, replace any bracketed expressions (which may be empty) with additional modifiers from the context or general knowledge that make the sentence more explicit. If there are no bracketed expressions or if there is no need to make any changes, do nothing. Do not change any content except for replacing brackets. \\
    \bottomrule
\end{tabular}%
}
\caption{The LLM prompts that are used for replacing at each edit node.}
\label{tab:prompts-replace}
\end{table}

\section{Experiment Details}
\label{app:exp-details}

\subsection{Decontextualization Experiment Details}

\subsubsection{Subselection Procedure}

We use the same procedure for subselecting data to use for our generation experiments as \cite{choi-2021-decontextualization} for both the Decontext and Switchboard datasets, which we reproduce here. First, we remove all examples where three or more annotators (out of five in the case of Decontext; out of three in the case of Switchboard) marked decontextualization as ``impossible'', and then discard any remaining annotations that mark the example as ``impossible''. To select the human annotation for comparison, we sort the annotations by length (in raw bytes) in descending order, take the median output as the human annotation, and use the remaining annotations as our gold references for the automatic evaluation.

\subsubsection{Expert Evaluation Setup}

To collect expert annotations of decontextualization validity and preference, we randomly sample 100 items from the subset for which we've generated decontextualized sentences. For each item, we randomly swap the order of the generated output and the human annotation.

Each annotator -- i.e., two of the authors -- annotated each item with the following:

\begin{enumerate}
    \item Is candidate 1 a valid decontextualization? (I.e., is it in a form that can be understood without context?) Answer with ``y'' or ``n''.
    \item Is candidate 2 a valid decontextualization? Likewise, answer with ``y'' or ``n''.
    \item What is your preference between candidate 1 and candidate 2? Answer with ``1'' if candidate 1 is better, ``2'' if gist 2 is better, or ``0'' if you are indifferent between the two.
\end{enumerate}

\subsection{Switchboard Annotation Details}

Since the Switchboard corpus is a fairly noisy dataset containing annotated transcriptions of telephone conversations, we first clean the data using \textsc{gpt-3.5-turbo}. We split each full conversation into chunks of utterances such that each chunk fits within the LLM token limit, and generate cleaned conversations using the following prompt: \textit{``Given an annotated conversation between two people, clean the conversation by removing all annotations and backchannels.''}. We then re-combine the cleaned blocks and split each turn in the conversation into multiple sentences. Finally, we create sentence and context pairs using a sliding window of size 5 over each conversation.

After preprocessing the Switchboard data, we filter all sentences and keep those that have at least one turn of context, and that have at least 6 words. For each item, we remove all context turns except for the 2 most recent turns. Since there are many sentences in the dataset that cannot be decontextualized with only the 2 most recent turns, we also use \textsc{gpt-3.5-turbo} to rank items by their quality, according to the following prompt:

\begin{quote}
Given a sentence and context sentences, provide a numerical quality rating between 1 (worst quality) and 5 (best quality) based on the following criteria:

- Whether every sentence is fluent and natural.

- Whether the sentences have interesting content.

- Whether the sentence can be understood given the provided context.

Do not give an explanation. Just give a single integer between 1 and 5.
\end{quote}

We filter out all items that have a rating of less than 3, and then randomly select 150 of the remaining examples to annotate.

\begin{figure*}[h]
    \centering
    \includegraphics[width=\linewidth]{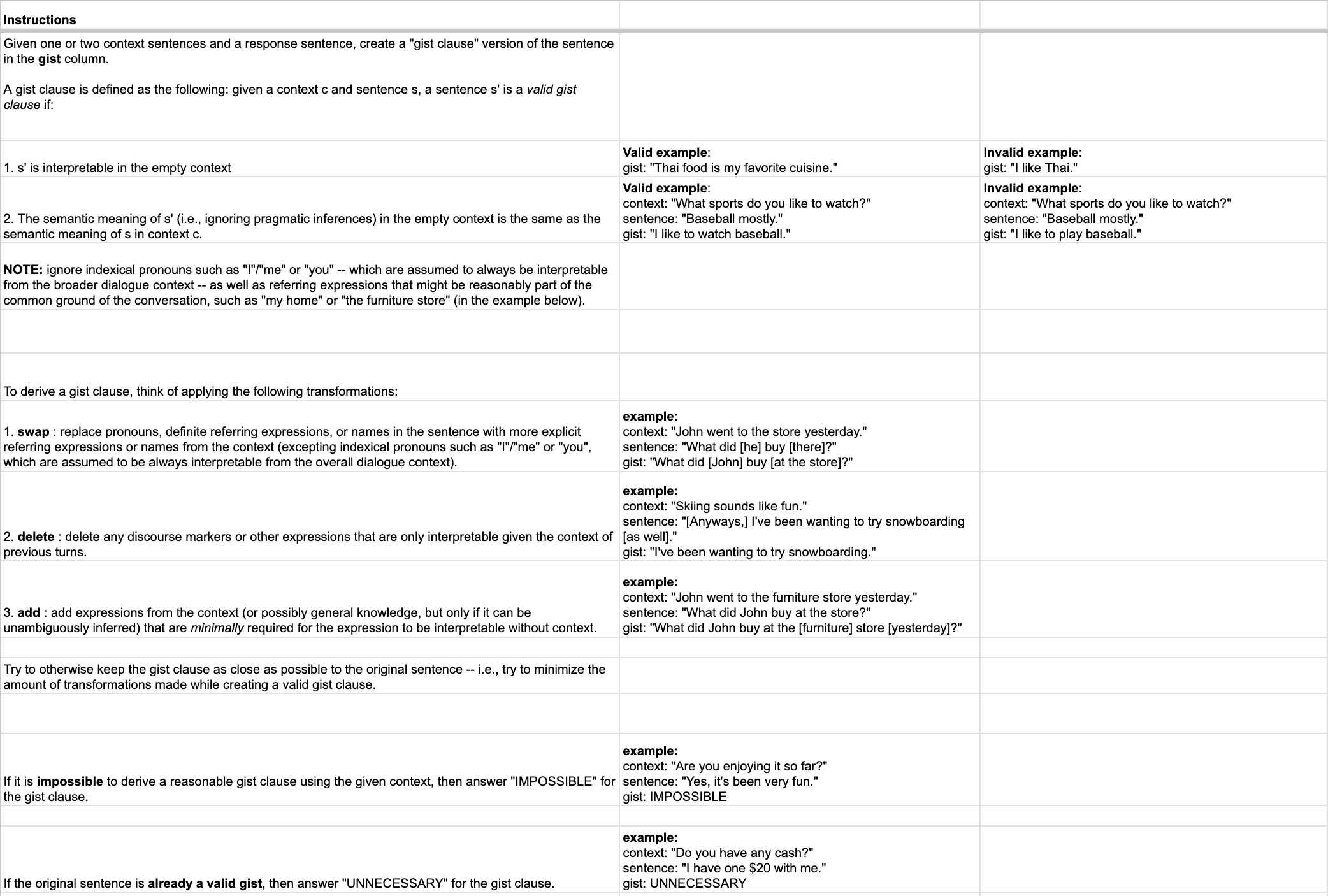}
    \caption{The instructions shown to annotators for annotating decontextualized sentences, or ``gist clauses'', in the Switchboard corpus.}
    \label{fig:switchboard-annotation}
\end{figure*}

The annotators (two of the authors, and one PhD student studying NLP in the same department) annotated each item with decontextualized sentences (referred to in the instructions as ``gist clauses'', due to the conversational setting) -- the instructions provided to annotators are shown in Figure \ref{fig:switchboard-annotation}.

\subsection{Experiment Costs}

We estimate that a full generation pass through $1000$ examples from the Decontext dataset, using $20$ examples for each edit step, cost about \$20 and took about 4 hours to complete. A full generation pass through $150$ annotated examples from the Switchboard dataset cost about \$2.5, and took about 30 minutes to complete.

\section{Examples}
\label{app:examples}

\subsection{Decontext dataset}

\begin{table}[th]
\resizebox{\columnwidth}{!}{%
\begin{tabular}{l | p{\linewidth}}
    \toprule
    \textbf{Context}   & A starting pistol or starter pistol is a blank handgun that is fired to start track and field races , as well as competitive swimming races at some meets . \\ \hline
    \textbf{Sentence}  & Beside sporting events , they are also used in films and in TV or stage shows . \\ \hline
    \textbf{LLM}       & Beside sporting events , starting pistols are also used in films and in TV or stage shows . \\ \hline
    \textbf{Human}     & Beside sporting events , starting pistols are also used in films and in TV or stage shows . \\
    \bottomrule
\end{tabular}%
}
\caption{A ``good'' example from the Decontext dataset (exact match)}
\label{tab:decontext-examples1}
\end{table}

\begin{table}[th]
\resizebox{\columnwidth}{!}{%
\begin{tabular}{l | p{\linewidth}}
    \toprule
    \textbf{Context}   & Geronimo Stilton is an Italian children 's book series published by Edizioni Piemme of Milan , Italy , since 2000 . \\ \hline
    \textbf{Sentence}  & Although the series credits its title character as the author , the idea was originally thought of by Elisabetta Dami . \\ \hline
    \textbf{LLM}       & Although the children ' s book series " Geronimo Stilton " credits its title character as the author , the idea was originally thought of by Elisabetta Dami , who is the actual author of the series . \\ \hline
    \textbf{Human}     & Although the Geronimo Stilton Italian children 's book series credits its title character as the author , the idea was originally thought of by Elisabetta Dami . \\
    \bottomrule
\end{tabular}%
}
\caption{A ``good'' example from the Decontext dataset}
\label{tab:decontext-examples2}
\end{table}

\begin{table}[th]
\resizebox{\columnwidth}{!}{%
\begin{tabular}{l | p{\linewidth}}
    \toprule
    \textbf{Context}   & ``Ms. Jackson'' is a song by American alternative hip hop duo OutKast . \\ \hline
    \textbf{Sentence}  & It was released on October 3 , 2000 , as the second single from their fourth album , Stankonia . \\ \hline
    \textbf{LLM}       & \textcolor{darkpastelred}{It} was released on October 3 , 2000 , as the second single of OutKast ' s fourth album ``Stankonia'' . \\ \hline
    \textbf{Human}     & \textcolor{darkpastelred}{The song ``Ms. Jackson''} was released on October 3 , 2000 , as the second single from the group OutKast fourth album , Stankonia . \\
    \bottomrule
\end{tabular}%
}
\caption{A ``bad'' example from the Decontext dataset (missing highlighted NP edit)}
\label{tab:decontext-examples3}
\end{table}

\begin{table}[th]
\resizebox{\columnwidth}{!}{%
\begin{tabular}{l | p{\linewidth}}
    \toprule
    \textbf{Context}   & On 8 May 1895 , the first North American cat show was hosted at Madison Square Garden in New York City. A female Maine Coon brown tabby , named Cosey , was entered into the show . \\ \hline
    \textbf{Sentence}  & Owned by Mrs. Fred Brown , Cosey won the silver collar and medal and was named Best in Show . \\ \hline
    \textbf{LLM}       & Owned by Mrs . Fred Brown , the female Maine Coon brown tabby named Cosey won the silver collar and medal and was named Best in Show . \\ \hline
    \textbf{Human}     & Owned by Mrs. Fred Brown , Cosey won the silver collar and medal and was named Best in Show \textcolor{darkpastelred}{in the first North American cat show in May 1895} . \\
    \bottomrule
\end{tabular}%
}
\caption{A ``bad'' example from the Decontext dataset (missing highlighted ADD edit)}
\label{tab:decontext-examples4}
\end{table}

In Tables \ref{tab:decontext-examples1}, \ref{tab:decontext-examples2}, \ref{tab:decontext-examples3}, and \ref{tab:decontext-examples4}, we provide some concrete examples of decontextualizations generated using our method for the Decontext dataset. We sample two examples of successful decontextualizations, and two examples of unsuccessful decontextualizations.

\subsection{Switchboard dataset}

In Tables \ref{tab:switchboard-examples1}, \ref{tab:switchboard-examples2}, \ref{tab:switchboard-examples3}, and \ref{tab:switchboard-examples4}, we provide some concrete examples of decontextualizations generated using our method for the Decontext dataset. We sample two examples of successful decontextualizations, and two examples of unsuccessful decontextualizations.

\begin{table}[th]
\resizebox{\columnwidth}{!}{%
\begin{tabular}{l | p{\linewidth}}
    \toprule
    \textbf{Context}   & What do you think, next to salary do you rank, is the most important job benefit? \\ \hline
    \textbf{Sentence}  & I would say the medical group insurance area as far as covering medical costs for yourself or family members. \\ \hline
    \textbf{LLM}       & I would say medical group insurance , as far as covering medical costs for yourself or family members , is the most important job benefit after salary . \\ \hline
    \textbf{Human}     & The medical group insurance area is the most important job benefit as far as covering medical costs for yourself or family members. \\
    \bottomrule
\end{tabular}%
}
\caption{A ``good'' example from the Switchboard dataset}
\label{tab:switchboard-examples1}
\end{table}

\begin{table}[t]
\resizebox{\columnwidth}{!}{%
\begin{tabular}{l | p{\linewidth}}
    \toprule
    \textbf{Context}   & I need to know how do you feel about sending an elderly family member to a nursing home? \\ \hline
    \textbf{Sentence}  & Of course, it's one of the last few things in the world you'd ever want to do, unless it's really for their own good. \\ \hline
    \textbf{LLM}       & Of course , the act of sending an elderly family member to a nursing home is considered one of the last resort options , unless it is genuinely in their best interest . \\ \hline
    \textbf{Human}     & You wouldn't want to send an elderly family member to a nursing home, unless you're sending them to the nursing home for thier own good. \\
    \bottomrule
\end{tabular}%
}
\caption{A ``good'' example from the Switchboard dataset}
\label{tab:switchboard-examples2}
\end{table}

\begin{table}[t]
\resizebox{\columnwidth}{!}{%
\begin{tabular}{l | p{\linewidth}}
    \toprule
    \textbf{Context}   & A: What kind of hobbies do you have? B: I do a lot of cross-stitching and painting, when I do have spare time. \\ \hline
    \textbf{Sentence}  & A: Really, I like cross-stitch too. \\ \hline
    \textbf{LLM}       & A: \textcolor{darkpastelred}{Really !} I like cross - stitch \textcolor{darkpastelred}{too} . \\ \hline
    \textbf{Human}     & A: I like cross-stitching. \\
    \bottomrule
\end{tabular}%
}
\caption{A ``bad'' example from the Switchboard dataset (missing highlighted DEL edits)}
\label{tab:switchboard-examples3}
\end{table}

\begin{table}[t]
\resizebox{\columnwidth}{!}{%
\begin{tabular}{l | p{\linewidth}}
    \toprule
    \textbf{Context}   & A: Okay, the topic was, should the United States adopt the metric system, and if so, should. \\ \hline
    \textbf{Sentence}  & B: How should they do it? \\ \hline
    \textbf{LLM}       & B: How should \textcolor{darkpastelred}{they} do \textcolor{darkpastelred}{it}? \\ \hline
    \textbf{Human}     & B: How should \textcolor{darkpastelred}{the United States} adopt \textcolor{darkpastelred}{the metric system}? \\
    \bottomrule
\end{tabular}%
}
\caption{A ``bad'' example from the Switchboard dataset (missing highlighted NP edits)}
\label{tab:switchboard-examples4}
\end{table}

\end{document}